\documentclass[conference]{IEEEtran}
\IEEEoverridecommandlockouts
\usepackage{cite}
\usepackage{cuted}
\usepackage{amsmath,amssymb,amsfonts}
\usepackage{algorithmic}
\usepackage{graphicx}
\usepackage{subfig}
\usepackage{textcomp}
\usepackage{xcolor}
\def\BibTeX{{\rm B\kern-.05em{\sc i\kern-.025em b}\kern-.08em
    T\kern-.1667em\lower.7ex\hbox{E}\kern-.125emX}}
\begin{document}

\title{Transformer-based Heuristic for Advanced Air Mobility Planning \\
}

\author{\IEEEauthorblockN{Jun Xiang}
\IEEEauthorblockA{\textit{Department of Aerospace Engineering} \\
\textit{San Diego State University}\\
San Diego, United State \\
jxiang9143@sdsu.edu}
\and
\IEEEauthorblockN{ Jun Chen}
\IEEEauthorblockA{\textit{Department of Aerospace Engineering} \\
\textit{San Diego State University}\\
San Diego, United State \\
Jun.Chen@sdsu.edu}

}

\maketitle

\begin{abstract}
Safety is extremely important for urban flights of autonomous Unmanned Aerial Vehicles (UAVs). Risk-aware path planning is one of the most effective methods to guarantee the safety of UAVs. This type of planning can be represented as a Constrained Shortest Path (CSP) problem, which seeks to find the shortest route that meets a predefined safety constraint. Solving CSP problems is NP-hard, presenting significant computational challenges. Although traditional methods can accurately solve CSP problems, they tend to be very slow. Previously, we introduced an additional safety dimension to the traditional A* algorithm, known as ASD A*, to effectively handle Constrained Shortest Path (CSP) problems. Then, we developed a custom learning-based heuristic using transformer-based neural networks, which significantly reduced computational load and enhanced the performance of the ASD A* algorithm. In this paper, we expand our dataset to include more risk maps and tasks, improve the proposed model, and increase its performance. We also introduce a new heuristic strategy and a novel neural network, which enhance the overall effectiveness of our approach.

\end{abstract}

\section{Introduction}\label{sec: I}

Path planning is an important task in many research areas, including aerospace. In recent years, extensive scholarly research has been dedicated to optimizing flight path planning for unmanned aerial vehicles (UAVs)~\cite{guo2021safety,schopferer2020minimum}. This research focuses on identifying optimal flight trajectories characterized by minimized operational costs and avoidance of high-risk zones. This issue can be formulated as a Constrained Shortest Path (CSP) problem. According to the previous research~\cite{wang1996quality}, the CSP problem is NP-hard even for acyclic networks.

Currently, the CSP problem can be addressed using various methods. One of the earliest approaches involves dynamic programming, specifically node labeling \cite{desrochers1988generalized}. Numerous dynamic programming-based methods have been proposed~\cite{marinakis2017hybrid, resende2014grasp, tilk2017asymmetry, wu2020probabilistically, song2023modeling}. Another significant approach is Lagrangian Relaxation, which relaxes the weight constraint \cite{handler1980dual}. To close the duality gap resulting from Lagrangian relaxation, the Kth-shortest path method is employed \cite{dumitrescu2003improved, eppstein1998finding}. Despite acceleration algorithms, these methods remain computationally expensive, often requiring days to solve complex problems. To accelerate computational processes, heuristic approaches and approximate algorithms have been developed \cite{xiang2023hybrid}. Heuristic methods prioritize speed in practical scenarios but can not guarantee solution quality. $\epsilon$-approximation algorithms ensure solutions within $(1 + \epsilon)$ times the optimal cost but work slowly due to rigorous quality assurance \cite{xiao2005constrained}. RRT*\cite{karaman2011incremental} can find the optimal path while handling constraints. However, RRT* requires extensive sampling and frequent rewiring, making it computationally expensive.

Graph searching is another popular method for solving the path planning of drones. In fully visible airspace, searching for the shortest path from a start node to a goal node is easy. Nevertheless, path planning of drones considers not only the distance of the path but also many other features such as efficiency, schedule, and most importantly safety. In real airspace, we need a large number of nodes and edges to model airspace. Therefore, a faster solver is urgently needed. D*~\cite{stentz1995focussed} and D* based method~\cite{likhachev2005anytime} were proposed to quickly find a suboptimal path and then continue searching for the optimal path. Although these methods are useful in practice, D* does not find the optimal path more quickly. Jump point search~\cite{harabor2016optimal} can accelerate the A* algorithm by skipping intermediate nodes that do not need to be explicitly considered. However, in the airspace, it is not safe to skip any node.

Deep learning has emerged as a prominent research area driven by stronger computing power and the rising demand for artificial intelligence.
One recent example of the success of deep learning is the Generative Pre-trained Transformer (GPT), which generates high-quality text, images, and even music. GPT proposed by OpenAI is based on the Transformer architecture~\cite{vaswani2017attention}. The Transformer architecture is designed to handle sequences of data, such as text or speech, and is well suited for natural language processing tasks. The latest research~\cite{dosovitskiy2020image, chen2024bridging, hu2023planning, yukun2019deep} also has proven the effectiveness of transformers in addressing computer vision-related problems. There are studies~\cite{lehnert2024beyond, reed2022generalist} utilize transformers directly to solve planning problems, yet these methods cannot guarantee to find valid solutions.

\begin{figure*}[ht]
    \centering
    \includegraphics[width=0.78\textwidth]{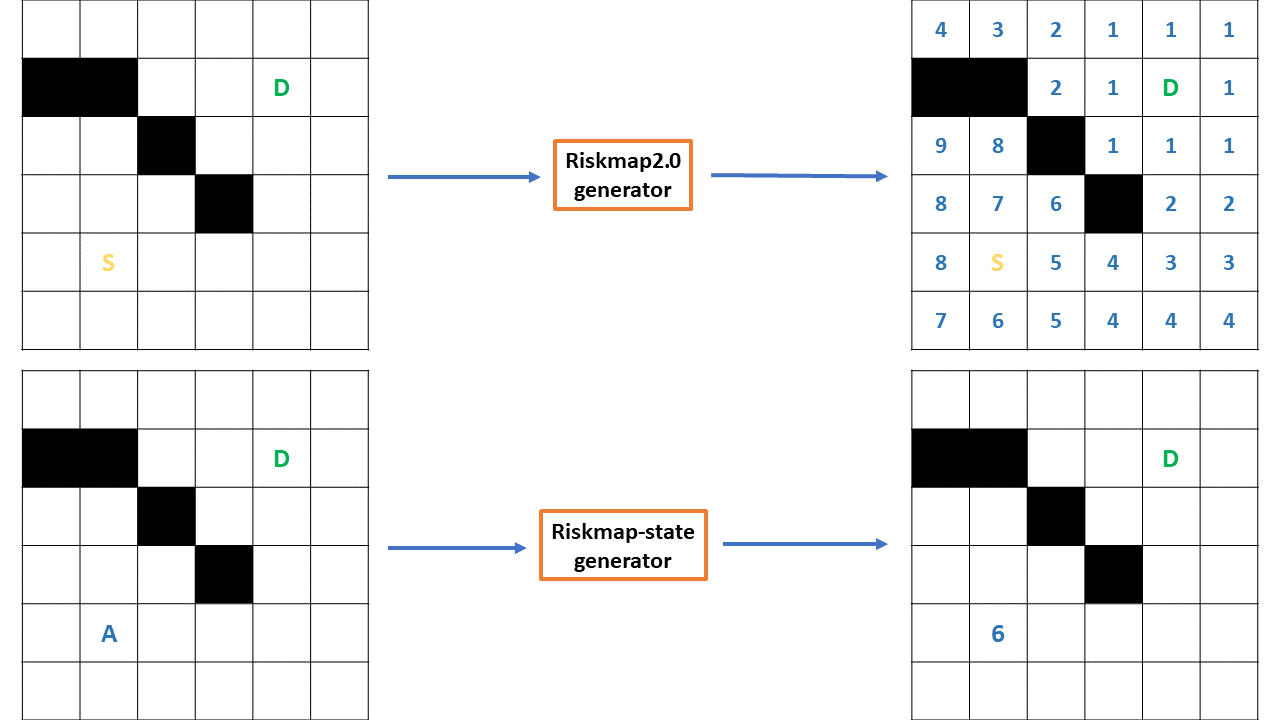}
    \caption{An example of heuristic generators: a) Riskmap2.0 takes the given start position (orange letter S), destination (green letter D), and risk map (black grid means the grid is blocked), generates heuristic (blue number) for all the available grid. b) Riskmap-state takes the current node (Blue letter A), destination, and risk map, generates heuristic (blue number) for the current node}
    \label{fig:introductionexample}
\end{figure*}
In this paper, we proposed transformer-based neural networks that can generate a good heuristic for the additional safety dimension A* (ASD A*)\cite{xiang2024learning} algorithm. With this neural network heuristic generator, the ASD A* algorithm can find the shortest path while exploring as less nodes as possible to accelerate the Advanced Air Mobility (AAM) planning. 

The main contributions are summarized as follows:
\begin{itemize}
    \item As shown in Fig.~\ref{fig:introductionexample}, we created two AAM planning datasets. One data set, called Riskmap2.0, includes the heuristics for all available grids given the start, destination, and risk map. This dataset basically contains the solution for the task.
    The other one, called Riskmap-state, records the heuristics for the current nodes given the current node, the destination, and the risk map. Riskmap-state focuses on the heuristic value of the current node instead of the task. It will be easier to learn. 
    
    \item  We introduce two transformer-based neural networks designed to generate heuristics for ASD A* to solve the AAM planning problem for the two datasets. The first neural network, trained on the Riskmap 2.0 dataset, generates heuristics for all grids in a risk map given the start position, destination, and risk map. The second neural network, trained on the Riskmap-State dataset, generates a heuristic only for a chosen (or current) node. 
    
    
    \item We conducted experiments on the heuristics generated by the proposed neural networks. The heuristics produced by the Riskmap 2.0 neural network outperformed the baseline heuristics on all 16*16, 24*24, and 32*32 maps. Additionally, experimental results indicate that the Riskmap-State neural network can generate heuristics close to expert-level heuristics, significantly accelerating the ASD A* algorithm. Furthermore, both neural networks demonstrate great potential and adaptability for risk maps with different sizes.

\end{itemize}

The rest of this paper is organized as follows. Section~\ref{sec:ps}
introduces the AAM planning problem and the special A* algorithm can solve the problem. Section~\ref{sec: dataset} describes the dataset generation process and specifies the attributes included in each data entry. Section~\ref{sec: NNs} introduces the two proposed neural network architectures in detail. Section~\ref{sec: results} shows
the experiment results. Finally, Section~\ref{sec: conclusion}
summarizes our conclusions and proposes the future development plan.

\begin{figure*}[htbp]
    \centering
    \subfloat[Risk map example]{\includegraphics[width=0.44\textwidth]{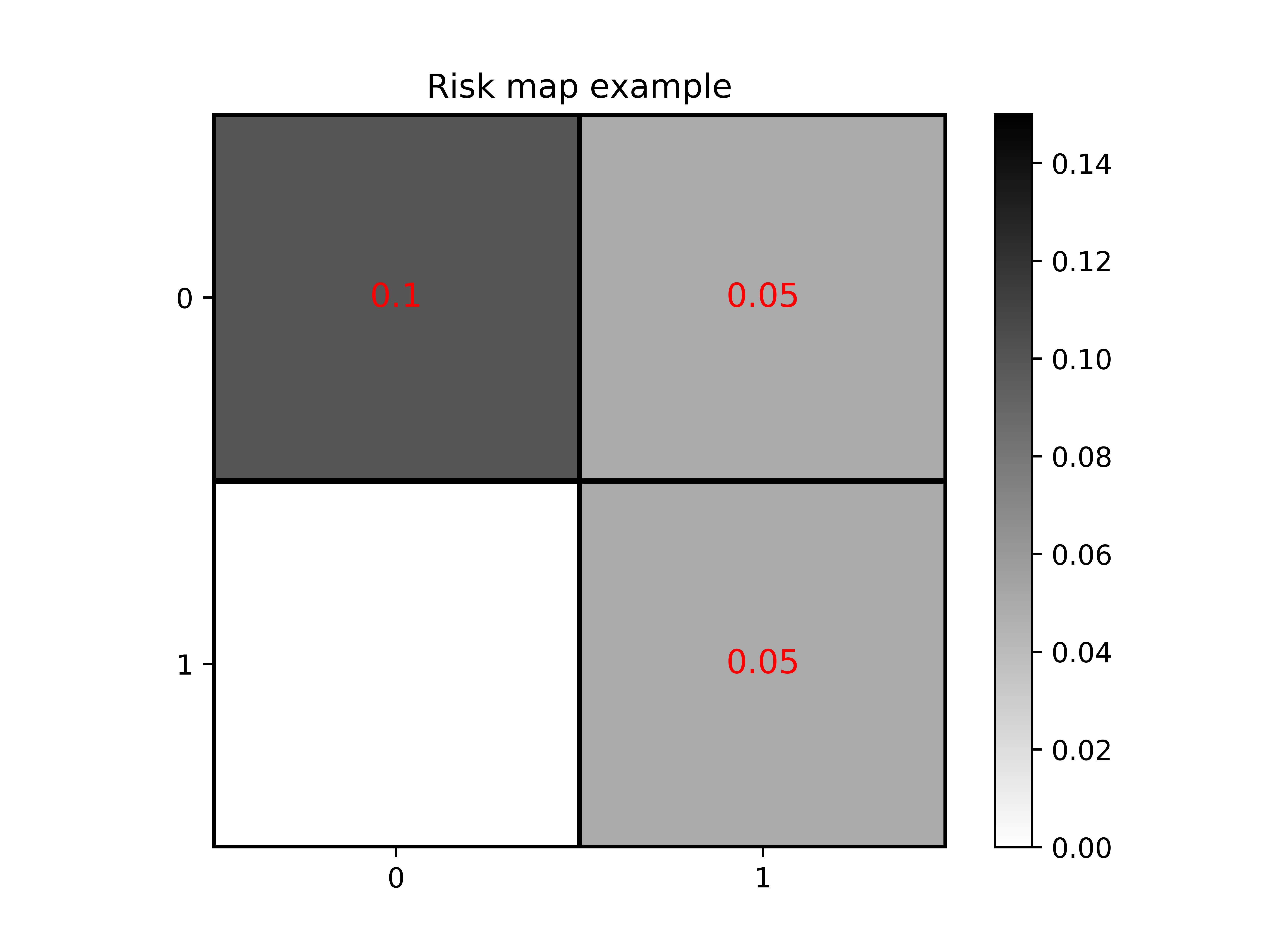}\label{fig:plot1}}
    \hfill
    \subfloat[Two possible path from grid (1, 0) to grid (0, 1)]{\includegraphics[width=0.44\textwidth]{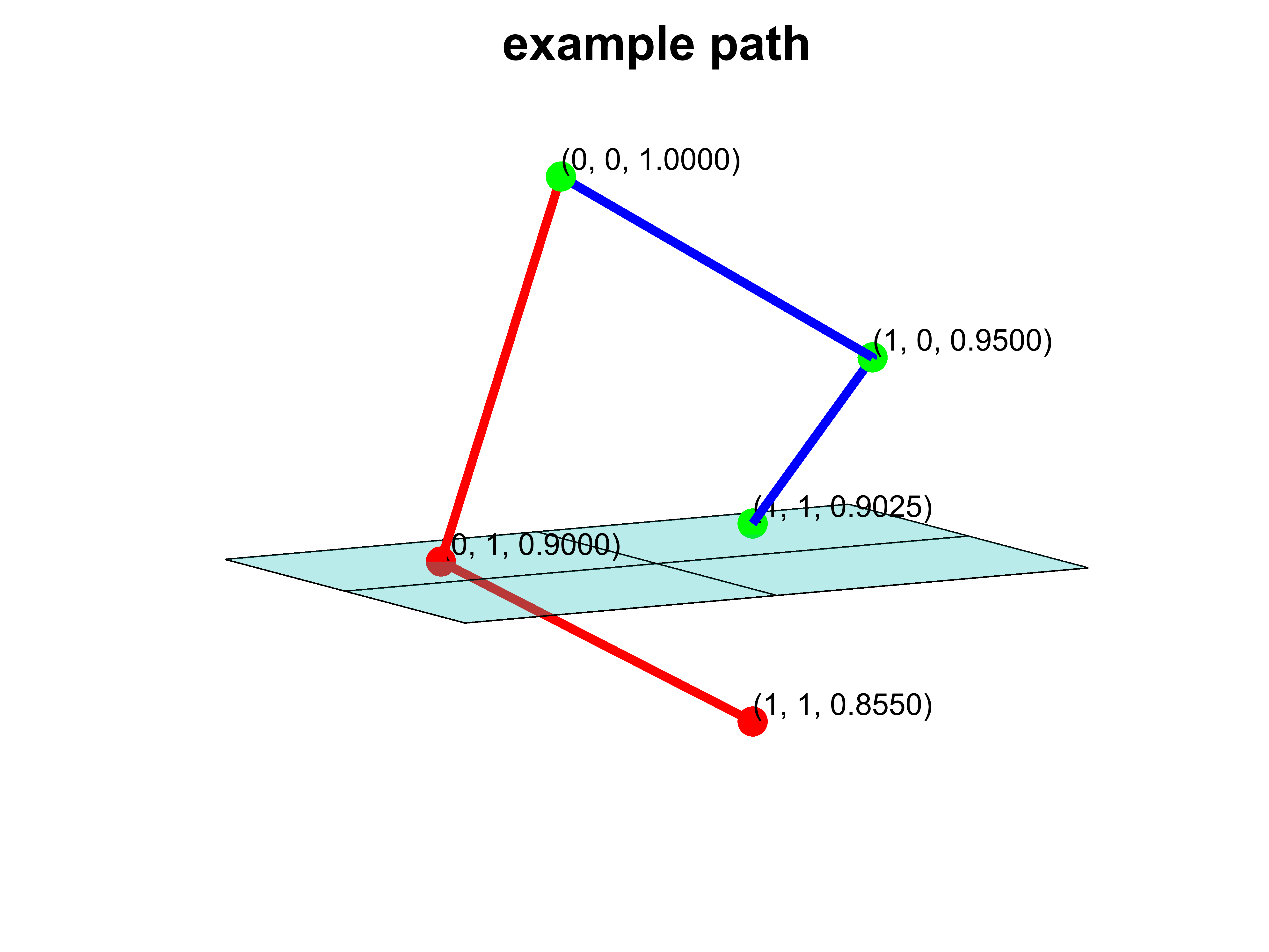}\label{fig:asd}}
    \caption{CSP example(2*2): The risk score of the grid (0, 0) is 0.1, the risk score of the grid (0, 1) and the grid (1, 1) is 0.05. There is a safety boundary(cyan plane) on the safety dimension, any path cross the safety boundary is invalid.}
    \label{fig: CSPexample}
\end{figure*}

\section{Problem Statement}\label{sec:ps} 
In this paper, we want to generate heuristics for ASD A*~\cite{xiang2024learning} to solve the AAM planning problem. The AAM planning problem we are trying to solve will be a CSP problem, introduced in our previous work\cite{xiang2024learning}. The ultimate goal is to find the risk-constrained shortest path on a risk map of airspace.
The CSP is formulated as follows:
\begin{align}
& \min \sum_{u, v \in N} c_{u v} z_{u v} \label{eq:4.1}\\
& \text { s.t. } \nonumber\\
& \sum_{\{v \mid u, v \in G\}} z_{u v}-\sum_{\{v \mid u, v \in G\}} z_{v u}= \begin{cases}1 & \text { for } u=g^s \\
-1 & \text { for } u=g^t  \\
0 & \text { otherwise }\end{cases} \label{eq:4.2}\\
& \prod_{z_{u v} = 1} S(v) \geq \epsilon \label{eq:4.3}\\
& \sum_{u, v \in G_s} z_{u v} \leq|G_s|-1 \text { for all } G_s \subseteq G\backslash g^s \label{eq:4.4}
\end{align}

where $c_{u v}$ is the distance cost of moving from grid $u$ to grid $v$, $z_{u v}$ is a binary variable determining if the path passes $e_{u v}$,  $z_{u v}=1$ if the edge is being traversed, otherwise $z_{u v}=0$.
The objective~(\ref{eq:4.1}) is to minimize the cost along the path subject to three constraints. The first constraint~(\ref{eq:4.2}) enforces the number of edges leading towards a grid is equal to the number of edges that leave that grid. 
The second constraint~(\ref{eq:4.3}) is a safety constraint where the accumulated safety (assuming each step is independent) in the path must be greater than the minimum safety threshold $\epsilon$. The third constraint~(\ref{eq:4.4}) eliminates the sub-tour by limiting the maximum number of edges in the path to the total nodes. $G_s$ is a subset of $\mathrm{G}$ that includes all nodes except the starting node.

We assume the graph is an undirected graph, which means if the agent can move from grid A to grid B, then the agent can move from grid B to grid A. At the same time, the safety threshold $\epsilon$ is always 0.9.  We also assume the agent can only move straight up (y - 1), down (y + 1), left (x - 1), and right (x + 1) for each step. The cost of moving from one grid to an adjacent one is considered to be uniform.

Fig~\ref{fig: CSPexample} shows an example of the ADS A* solving the AAM problem.
The ASD A* algorithm starts from the node (0, 0, 1), where the first two dimensions are the position and the third is the safety level. In the first step, it finds two neighbor nodes (0, 1, 0.9) and  (1, 0, 0.95). By exploring these two nodes, the algorithm can find two paths. However, because the red path crosses the safety boundary, it is not valid. The only feasible path is the blue path in the figure.

\section{Training Dataset}\label{sec: dataset}
Each dataset entry includes a risk map, a destination (task), and the corresponding expert heuristic.
\subsection{Risk map and destination}
We create random risk maps of varying sizes. For the Riskmap 2.0 dataset, the sizes include 16*16, 24*24, and 32*32. For the Riskmap-State dataset, the sizes include 16*16 and 64*64. Each random map contains an equal number of low-risk, high-risk, and safe grids. Low-risk grids have a risk score ranging from 0 to 0.1, resulting in the agent accumulating a risk penalty when passing through them. High-risk grids have a risk score of 1 and should be avoided by the agent. Safe grids have a risk score of 0, allowing the agent to pass through without any risk penalty. We then randomly select a safe grid or a low-risk grid as the destination.

\subsection{Expert heuristic}
We introduced two heuristic-generating strategies. The first strategy generates a heuristic for each grid given the start position, destination, and risk map. We use this strategy to create the Riskmap2.0 dataset. The second strategy generates a heuristic for the current node based on the current node, destination, and risk map. We use this strategy to create the Riskmap-state dataset.
As mentioned in Sec~\ref{sec:ps}, all the heuristics are for the safety threshold $\epsilon$ equal to 0.9. The detailed process for creating the dataset is as follows:
\subsubsection{Riskmap2.0}
For each risk map and destination, we pick a random start position. Given the start position, destination, and risk map, we can use a traditional solver to find the shortest path. For all grids not on the shortest path, the expert heuristic will be the Manhattan distance from the grid to the destination plus a penalty. Conversely, the heuristic for the grids on the shortest path will be equal to the Manhattan distance. This approach leads to nodes on the shortest path having a lower heuristic while maintaining consistency. If ASD A* uses this heuristic for the corresponding grid to add a node during the search process, it will only explore the nodes on the shortest path, allowing it to find the shortest path very quickly. We have created more than 128,000 data entries for each map size. 

\subsubsection{Riskmap-state}
For each risk map with a fixed destination, we randomly select a grid and assign it a random safety level between 0.9 and 1 to construct a current node during the A* search. 
Then, we use a traditional solver to find the shortest path between the current node and the destination. Consequently, we use the searched shortest distance between the current node and the destination as its heuristic for the current node. Since this heuristic is exactly equal to the actual distance, the ASD A* algorithm can find the shortest path without exploring any nodes that are not on the shortest path. We have created more than 128,000 data entries for each map size.

\section{Neural networks}\label{sec: NNs}
\begin{figure*}[t]
    \centering
    \includegraphics[width=0.95\linewidth]{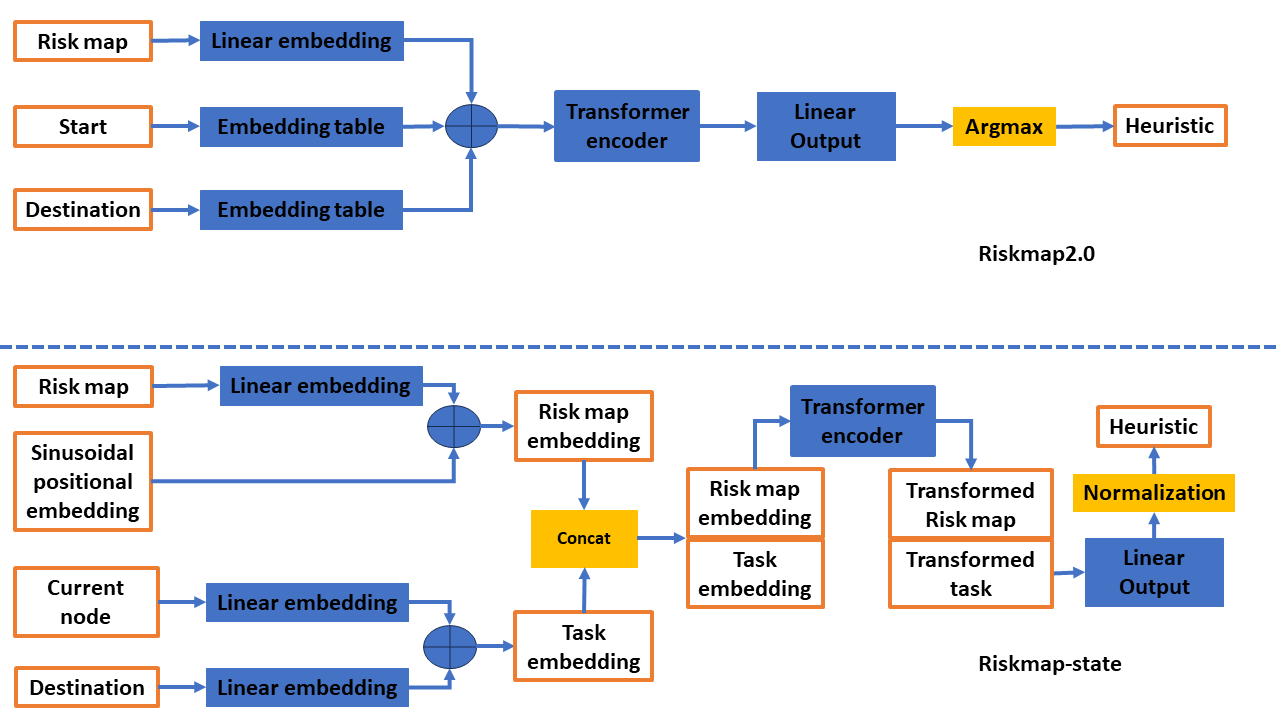}
    \caption{Proposed transformer heuristic architecture }
    \label{fig: architecute}
\end{figure*}

We proposed two transform-based heuristic-generating neural networks as shown in Fig.~\ref{fig: architecute}. The first is for the Riskmap2.0 dataset, which is an upgraded version of our previous work~\cite{xiang2024learning}. This neural network can take the risk map, the start position, and the destination as inputs and generate the heuristic for each grid in the risk map. In contrast, the second one, which we designed for the Riskmap-state dataset, can take the risk map, a node of the ASD A* process, and the destination as inputs and generate the heuristic of the input node. When training Riskmap2.0, we train the model with datasets of different map sizes separately. In contrast, we train Riskmap-state with datasets of different map sizes together.

\subsection{Overall architecture}
\subsubsection{Riskmap2.0}
\begin{figure}[t]
    \centering
    \includegraphics[width=\linewidth]{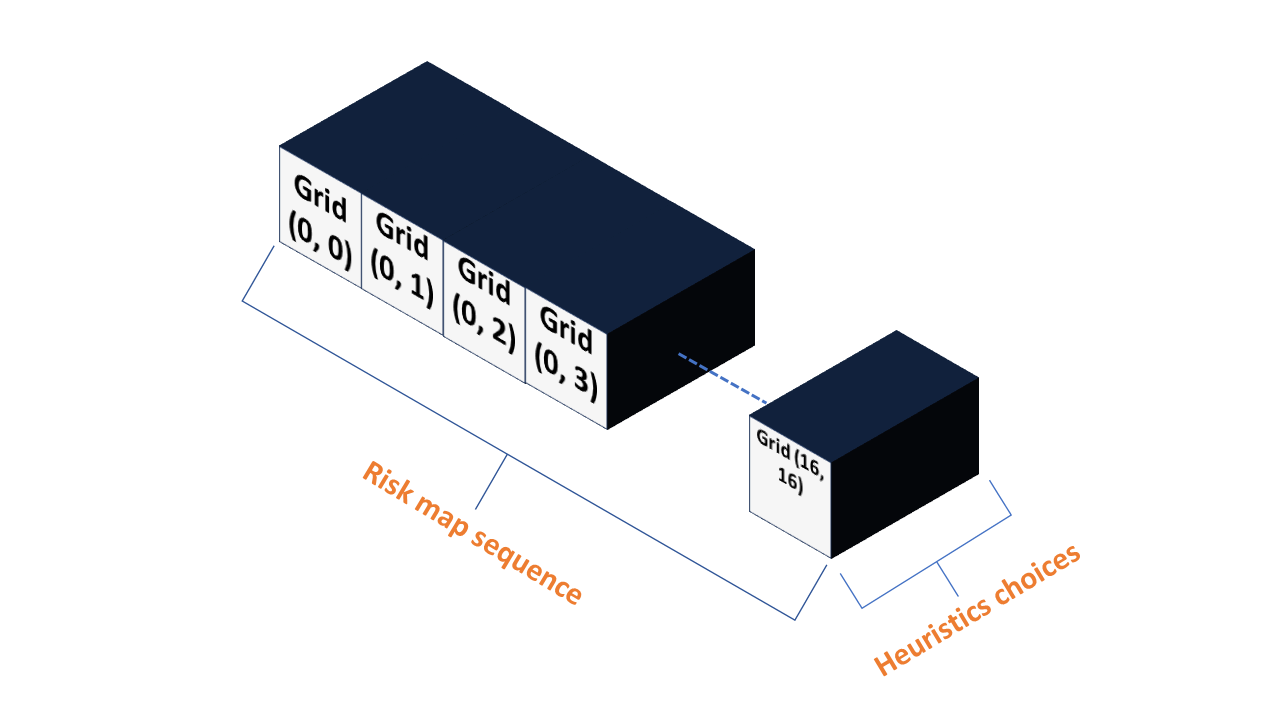}
    \caption{Output of the Riskmap2.0(16*16)}
    \label{fig: riskmap2.0output}
\end{figure}
The overall architecture of Riskmap 2.0 is shown in the upper part of Fig.~\ref{fig: architecute}. We tokenized the starting position and destination based on the map size. For example, on a 16*16 map, (0, 0) is tokenized to integer 0, and (1, 1) is tokenized to integer 17. In addition, we flatten the risk map into a 1D sequence. For instance, on a 24*24 map, the risk of the (0, 0) grid becomes the 0th element in the flattened risk map sequence, and the risk of the (1, 1) grid becomes the 25th element in the flattened risk map sequence. 
Then, we embed the tokenized start and destination using an embedding table. The embedding table maps the tokenized start and destination to their corresponding dense vector representations. On the other hand, we use a single-layer linear layer to embed the risk map sequence into the same dimension. Since we train the model on datasets with different map sizes separately, we removed the position embedding because the position of each sequence in the risk map is fixed. We duplicate the embeddings of the start and destination to match the length of the risk map sequence. Then, the feature of each grid in the risk map sequence is combined with the embeddings of the start and destination. The combined item will then be sent to the transformer. The output layer, as shown in Fig.~\ref{fig: riskmap2.0output} will decrease the dimension of the transformed item. Each value of each feature for each grid represents the probability that the heuristic of this grid is the corresponding feature. Finally, the Argmax function can choose the heuristic with the highest probability for all the grids. The advantage of this classification-style output is that the generated heuristic can be bound in a certain set of value. 

\subsubsection{Riskmap-state}
\begin{figure}[t]
    \centering
    \includegraphics[width=\linewidth]{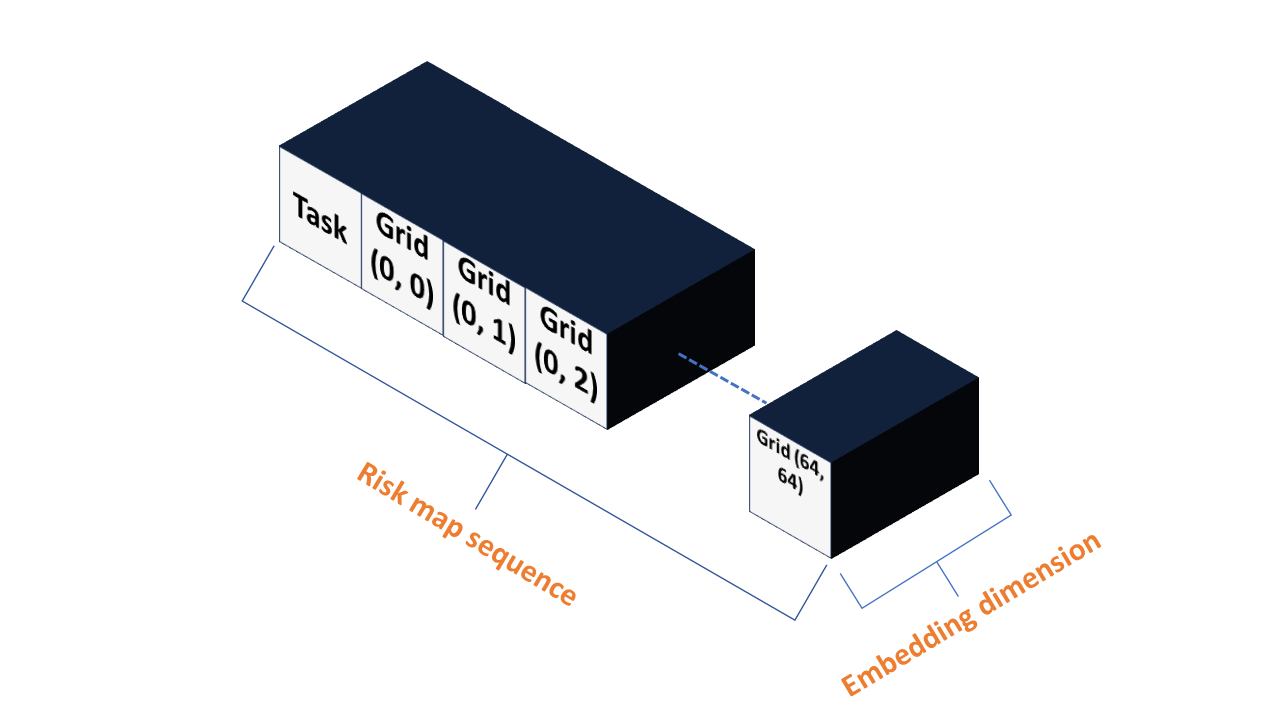}
    \caption{Concatenated embedding of the Riskmap-state(64*64)}
    \label{fig: riskmapstateEmbed}
\end{figure}
The overall architecture of Riskmap-State is shown in the lower part of Fig.~\ref{fig: architecute}. Similarly to Riskmap 2.0, we flatten the risk map into a 1D sequence and embed it using a single linear layer. The size of the risk map can vary, so we pad the risk map with unsafe grids to ensure uniform tensor dimensions for efficient batch processing.
We also combine the embedded risk map with sinusoidal positional embeddings \cite{vaswani2017attention}, allowing the transformer to understand the position of each grid. The current node, originally a 3D vector, and the destination, originally a 2D vector, are also embedded into the embedding dimension using a single linear layer. These are combined to form the task embedding. Then, we concatenate the risk map embedding and task embedding together. The concatenated embedding is illustrated in Fig.~\ref{fig: riskmapstateEmbed}.
After passing through the transformer encoder, we use a linear output layer to process the task embedding channel, which has interacted with all the grid channels during the transformer step. The final output is normalized to ensure it falls within a reasonable range for heuristic purposes.

\subsection{Transformer and loss function}
We use the original Transformer encoder \cite{vaswani2017attention} as the backbone for both of our neural networks. A mask is not required because if the map is smaller than the maximum map size, we treat the padding grids as unsafe grids. Thus, there is no issue with other grids interacting with the padding grids.

For Riskmap2.0, we use the cross-entropy loss function, which is widely used in classification tasks. The cross-entropy loss function~(\ref{eq: loss}) compares the predicted probability distribution over classes (heuristics in our case) with the true labels (expert heuristics).
\begin{equation}
L = -\frac{1}{N} \sum_{i=1}^N \sum_{j=1}^C y_{ij} \log(p_{ij}) \label{eq: loss}
\end{equation}

where:
\begin{itemize}
  \item \(N\) is the number of grids.
  \item \(C\) is the number of heuristic choices.
  \item \(y_{ij}\) is a binary indicator (0 or 1) if heuristic \(j\) is equal to the expert heuristic for the grid \(i\).
  \item \(p_{ij}\) is the predicted probability that the heuristic of the grid \(i\) is heuristic \(j\).
\end{itemize}

For Riskmap-state, we use the MSE loss function~(\ref{eq: mse}) as in many regression works. MSE loss can help the neural network minimize the average error between the expert and the output. 
\begin{equation}
\text{MSE} = \frac{1}{n} \sum_{i=1}^{n} (y_i - \hat{y}_i)^2
\label{eq: mse}
\end{equation}
where:
\begin{itemize}
    \item \(n\) is the number of task,
    \item \(y_i\) is the expert value for the \(i\)-th task,
    \item \(\hat{y}_i\) is the generated value for the \(i\)-th task.
\end{itemize}

\begin{figure*}[htbp]
    \centering
    \subfloat[Node explored by ASD A* with Manhattan heuristic]{\includegraphics[width=0.46\textwidth]{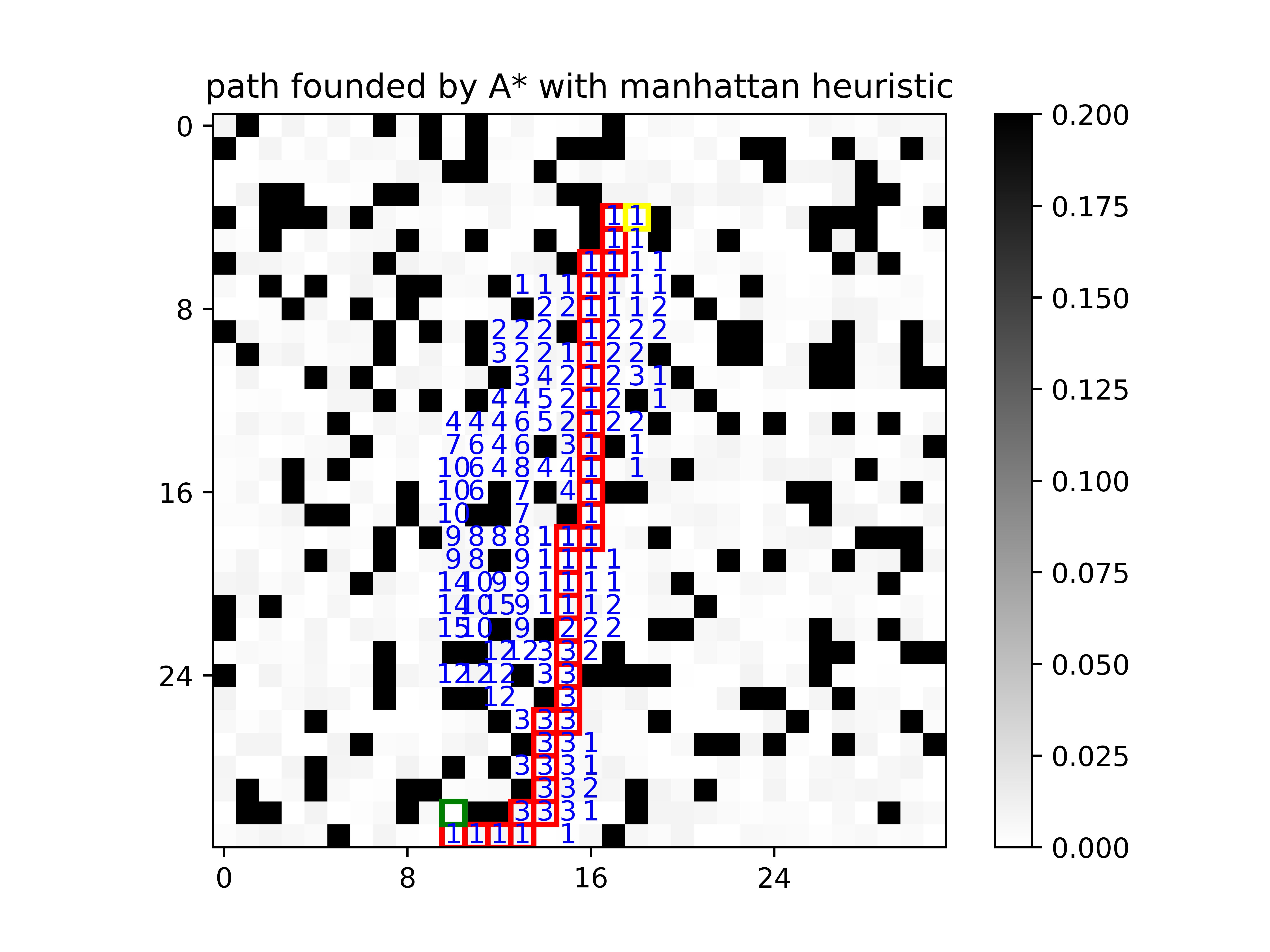}\label{fig:plot1}}
    \hfill
    \subfloat[Node explored by ASD A* with Riskmap2.0 heuristic]{\includegraphics[width=0.46\textwidth]{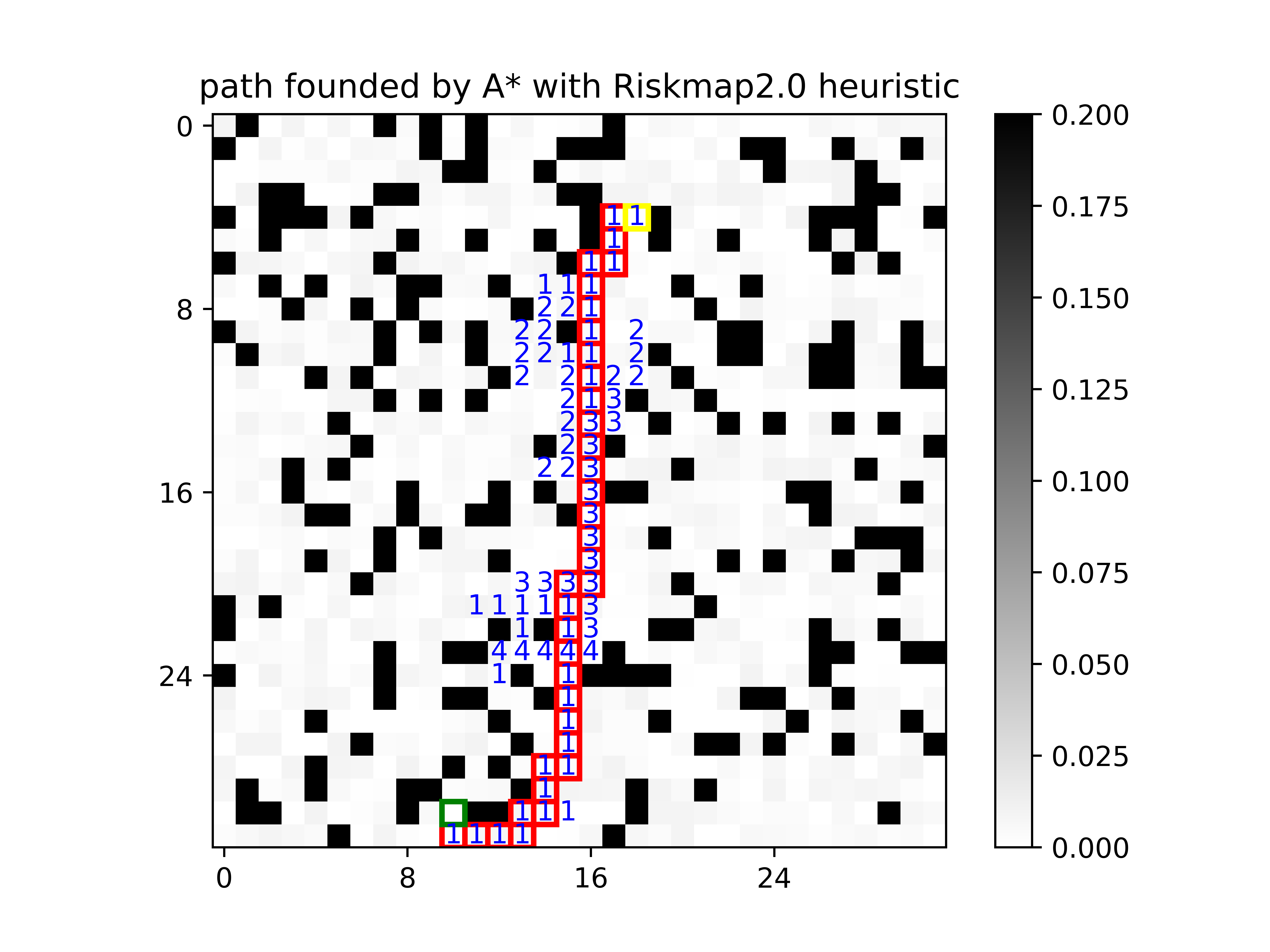}\label{fig:plot2}} 
    \caption{Result example(32*32 random): blue number means how many times a node is explored on this grid}
    \label{fig: riskmap2.0example}
\end{figure*}

\begin{table*}[h]
    \centering
    \caption{Riskmap2.0 heuristic vs Manhattan heuristic}
    \label{tab:riskmap2.0result}
    \begin{tabular}{lcccc}
    \hline  Mapsize & Method & Node explored($\downarrow$) & Search time(ms) & SPL ($\uparrow$) \\
    \hline \hline
    16*16&Riskmap2.0(our) & $35.38$ & 1.14 & $99.58\%$ \\ 
   & Manhanton & $58.48$ & 1.50& $100\%$  \\ 
   \hline
   24*24 &Riskmap2.0(our) & 214.96 & 6.33& $97.55\%$ \\
   & Manhanton & 258.18 & 7.61& $100\%$  \\
   \hline
    32*32&Riskmap2.0(our) & $754.19$ & 24.55& $98.27\%$ \\
   & Manhanton & $826.14$ & 27.19& $100\%$  \\
    
    \hline
    \end{tabular}
\end{table*}

\section{Result}\label{sec: results}
\subsection{Riskmap2.0}
We evaluate the generated heuristic with the average number of nodes explored, the average search time, and the SPL\cite{anderson2018evaluation}, where SPL stands for Success weighted by (normalized inverse) Path Length that is defined by the following equation:
\begin{equation}
\frac{1}{N} \sum_{i=1}^N S_i \frac{\ell_i}{\max \left(p_i, \ell_i\right)} .
\end{equation}
where:
\begin{itemize}
  \item \(N\): The total number of test episodes conducted.
  \item \(S_i\): A binary indicator of success in episode \(i\), where \(S_i = 1\) if the agent successfully reaches the goal, and \(S_i = 0\) otherwise.
  \item \(\ell_i\): The shortest path distance from the agent's starting position to the goal in episode \(i\).
  \item \(p_i\): The length of the path actually found by the algorithm in episode \(i\).
  \item \(\max(p_i, \ell_i)\): The maximum value between the shortest path distance and the actual path taken for episode \(i\), ensuring the denominator is at least as large as the shortest path.
\end{itemize}
Riskmap2.0 is tested in 16*16, 24*24, and 32*32 grid maps. For each map size, we randomly pick 1000 tasks from 100 unseen risk maps.  Fig. \ref{fig: riskmap2.0example} shows an example comparison between the Riskmap2.0 heuristic and the Manhattan heuristic. The results are shown in Table~\ref{tab:riskmap2.0result}. According to the result, With the heuristic Riskmap2.0 generated, the A* algorithm explores fewer nodes and needs less time to find the path to the destination in all three datasets. Especially for the 16*16 dataset, the number of explored nodes is $39.5\%$ less than the Manhattan, and the average search time is $24\%$ shorter while achieving a $99.58\%$ SPL score, indicating high optimality. For the 24*24 dataset, the number of nodes explored is $16.75\%$ less than the Manhattan, and the average search time is $16.82\%$ shorter while achieving a $97.55\%$ SPL score. For the 32*32 dataset, the number of nodes explored is $8.71\%$ less than the Manhattan, and the average search time is $9.71\%$ shorter while achieving a $98.27\%$ SPL score.

The results show that the RiskMap2.0 model can successfully accelerate the A* algorithm while maintaining very high optimality. However, performance decreases as the map size increases. This is because the input size of the neural network increases as the map size grows, leading to higher training costs. Additionally, data generation becomes slower, so we have to reduce the number of tasks per map in the training dataset. Consequently, with the same amount of time and computing resources, the number of training epochs and the quality of the data decrease as the map size increases.

\subsection{Riskmap-state}
The trained Riskmap-state model can generate a heuristic close to the expert heuristic. We test the Riskmap-state on 1000 random tasks from 100 unseen risk maps. The MSE loss for the 16*16 risk map is 1.742, and the MSE loss for the 64*64 risk map is 6.234. With the Riskmap-state heuristic, A* can also explore fewer nodes. For the 16*16 dataset, the number of nodes explored is $19.23\%$ less than the Manhattan. for the 16*16 dataset, the number of nodes explored is $10.51\%$ less than the Manhattan. However, if the map is small and the task is simple, the heuristic generating time may be longer than the search time, because the Riskmap-state must generate a heuristic for every node A* found. We also can not guarantee to find the optimal path with the Riskmap-state heuristic.
\begin{figure}[t]
    \centering
    \includegraphics[width=\linewidth]{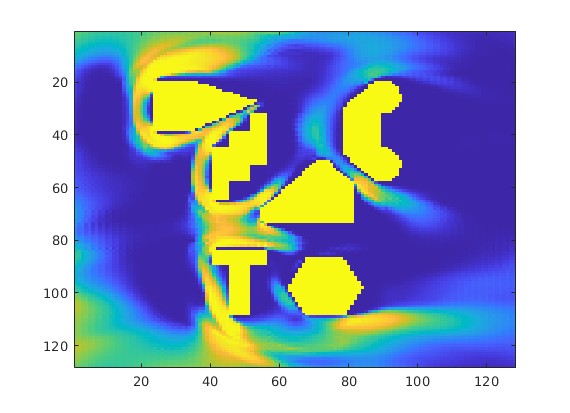}
    \caption{City risk map with wind flow near building}
    \label{fig: flowmap}
\end{figure}

\begin{figure*}[htbp]
    \centering
    
    \subfloat[Node explored by ASD A* with Manhattan heuristic in 16*16 city map]{\includegraphics[width=0.46\textwidth]{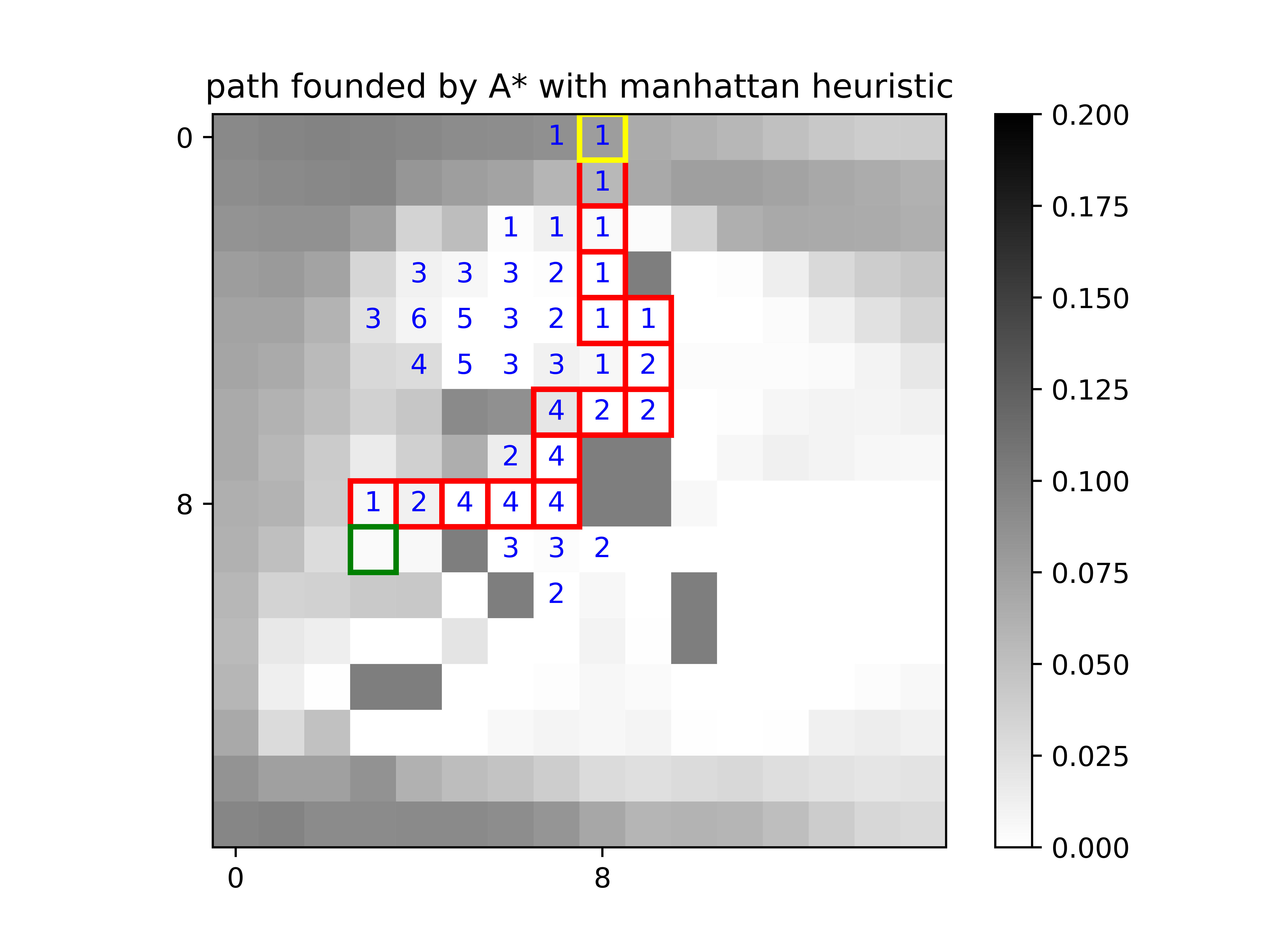}\label{fig:plot2}}
    \hfill
    \subfloat[Node explored by ASD A* with Riskmap2.0 heuristic in 16*16 city map]{\includegraphics[width=0.46\textwidth]{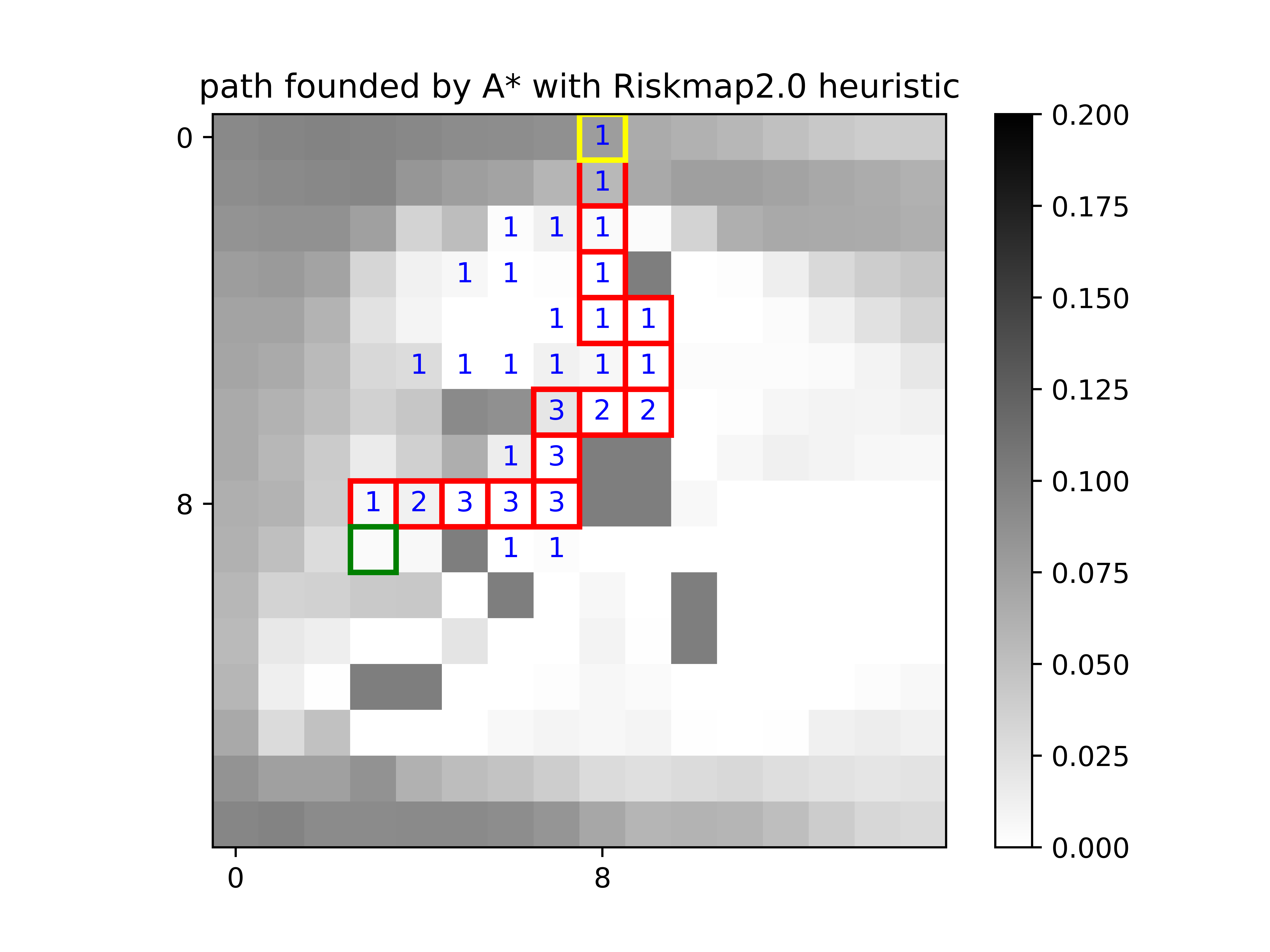}\label{fig:plot2}}
    \caption{Result example (16*16 city): blue number means how many times a node is explored on this grid}
    \label{fig: riskmap2.0realexample}
\end{figure*}
\subsection{Realistic city environment}
In addition to pure random maps, we also tested our method on a set of 16*16 risk maps, which we call a city map. The city map is based on wind flow near high buildings generated by a city wind flow simulator~\cite{gault2023safe}. The city wind flow simulator can calculate the risk of each grid turbulence given the position of the buildings, the wind speed, and the height of the assessment. Fig. \ref{fig: flowmap} shows an example of wind flow in a 128*128 map. We scale those 128*128 maps down to 16*16 maps by dividing the 128*128 maps into 256 8*8 grids and calculating the mean value for each 8 * 8 grid.  
We also tested 1000 tasks picked from 100 different risk maps. The average number of nodes explored (20.49) of our method is $53.53\%$ less than the Manhattan (44.09), and the average search time (0.38 ms) is $48.65\%$ shorter than the Manhattan (0.74 ms) while achieving a $99.73\%$ SPL score. Fig. \ref{fig: riskmap2.0realexample} shows an example of how Riskmap2.0 heuristic beat the Manhattan heuristic on this city map.

\section{Discussion and conclusion}\label{sec: conclusion}
The results have proven that the Riskmap2.0 model can accelerate the A*-like algorithm without losing too much optimality in a relatively small map. Additionally, the Riskmap-state also shows the ability to generate accurate expert heuristics. Riskmap2.0 can generate the solution directly for the A*. If heuristic generating is correct, the process can be very fast. However, if the heuristic generating is wrong, the heuristic may be very misleading. Riskmap-state can generate a heuristic for any nodes found during the A* process. A* can only explore the shortest path as well if the heuristic generating is correct. However, riskmap2.0 needs to be processed for every node A* process found.

Compared to state-of-the-art AI path planners, such as DeepCube~\cite{mcaleer2018solving}, Gato~\cite{reed2022generalist}, and searchformer~\cite{lehnert2024beyond}, the biggest advantage of our method is we can $100\%$ solve the problem while previous planning generating AI can not guarantee that. At the same time, we also achieve higher optimality in the complex task. Moreover, the proposed method requires less training time and a smaller dataset. Our model is also smaller and requires fewer computing resources during the operation.
Our research agrees with the previous research~\cite{lehnert2024beyond} that planning on a larger grid map is still challenging. In the future, we aim to decrease the model generation time and expand the dataset by creating more expert heuristics for larger risk maps.

\bibliographystyle{unsrt}
\bibliography{main}

\end{document}